%% file: paper.tex
\documentclass[11pt]{article}
\usepackage{colacl} 
\usepackage{psfig}
\usepackage{epsf}
\title{Processing Self Corrections in a speech to speech system}
\author{J{\"o}rg Spilker, Martin Klarner, G{\"u}nther G{\"o}rz\\
  University of Erlangen-Nuremberg - Computer Science Institute, \\ 
  IMMD 8 - Artificial Intelligence, \\
  Am Weichselgarten 9, 91058 Erlangen - Tennenlohe, Germany \\
  {\tt \{spilker,klarner,goerz\}@immd8.informatik.uni-erlangen.de}}

\begin{document}

\maketitle

\begin{abstract}
Speech repairs occur often in spontaneous spoken dialogues. The
ability to detect and correct those repairs is necessary for any
spoken language system. We present a framework to detect and correct
speech repairs where all relevant levels of information,
i.e., acoustics, lexis, syntax and semantics can be integrated. The
basic idea is to reduce the search space for repairs as soon as
possible by cascading filters that involve more and more
features. At first an acoustic module generates hypotheses about the
existence of a repair. Second a stochastic model suggests a correction
for every hypothesis. Well scored corrections are inserted as new
paths in the word lattice. Finally a lattice parser decides on accepting the repair. 
\end{abstract}

\bibliographystyle{acl}

\input {text1.tex}

\end{document}

%% file: text1.tex
\newcommand{\vmsp}{VERBMOBIL }
\newcommand{\vm}{VERBMOBIL}
\newlength{\itemspace}
\setlength{\itemspace}{-0.22cm}
\newlength{\eqspace}
\setlength{\eqspace}{-0.15cm}
\setlength{\eqspace}{-0cm}
\newlength{\eqendspace}
\setlength{\eqendspace}{-0.1cm}
\setlength{\eqendspace}{-0cm}
\newlength{\itspace}
\setlength{\itspace}{-0.1cm}
\setlength{\itspace}{-0cm}
\newlength{\itendspace}
\setlength{\itendspace}{-0.1cm}
\setlength{\itendspace}{-0.1cm}

\section{Introduction}

\begin{figure}[htb]
\begin{center}
\mbox{\epsfxsize=7cm
        \epsffile{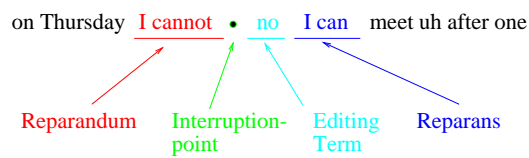}} 
\end{center}
\vspace*{-.65cm}
\caption{\label{segmentation}Example of a self repair}
\end{figure}

Spontaneous speech is disfluent. In contrast to read speech the
sentences aren't perfectly planned before they are uttered. Speakers
often modify their plans while they speak. This results in pauses,
word repetitions or changes, word fragments and restarts. Current
automatic speech understanding systems perform very well in small
domains with restricted speech but have great difficulties to deal
with such disfluencies. A system that copes with these self corrections
(=repairs) must recognize the spoken words and identify the repair to
get the intended meaning of an utterance. To characterize a repair it
is commonly segmented into the following four parts (cf. fig.\ref{segmentation}): 
\begin{itemize}
\item reparandum: the ``wrong'' part of the utterance  \vspace*{\itemspace}
\item interruption point (IP): marker at the end of the reparandum  \vspace*{\itemspace}
\item  editing term: special phrases, which indicate a repair like
``well'', ``I mean'' or filled pauses such as ``uhm'', ``uh'' 
\item reparans: the correction of the reparandum  \vspace*{\itemspace}
\end{itemize}
\vspace*{-.35cm}
\vspace*{-.2cm}
Only if reparandum and editing term are known, the utterance can be
analyzed in the right way. It remains an open question whether the
two terms should be deleted before a semantic analysis as suggested
sometimes in the literature\footnote{In most cases a reparandum could be deleted
without any loss of information. But, for example, if it introduces an
object which is referred to later, a deletion is not appropriate.}. If
both terms are marked it is a straightforward preprocessing step to delete
reparandum and editing term. In the Verbmobil\footnote{This work is part of the  
\vmsp project and was funded by the German Federal Ministry for Research and Technology
(BMBF) in the framework of the Verbmobil Project under Grant BMBF 01
IV 701 V0. The responsibility for the contents of this study lies
with the authors.} corpus, a corpus dealing with 
appointment scheduling and travel planning, nearly 21\% of all turns
contain at least one repair. As a consequence a speech understanding
system that cannot handle repairs will lose performance on these
turns.

Even if repairs are defined by syntactic and semantic
well-formedness \cite{levelt:1983} we observe that most of
them are local phenomena. At this point we have to differentiate
between restarts and other repairs\footnote{Often a third kind of repair
  is defined: ``abridged repairs''. These repairs consist solely of an
  editing term and are not  repairs in our sense.} (modification repairs).
Modification repairs have a strong correspondence between reparandum and reparans,
whereas restarts are less structured. In our believe there is no need
for a complete syntactic analysis to detect and correct most
modification repairs. Thus, in what follows, we will concentrate
on this kind of repair.

There are two major arguments to process repairs before parsing.
Primarily spontaneous speech is not always syntactically well-formed even
in the absence of self corrections. Second (Meta-) rules increase the parsers'
search space. This is perhaps acceptable for transliterated speech but
not for speech recognizers output like lattices because they represent
millions of possible spoken utterances.
In addition, systems which are not based on a deep syntactic and semantic
analysis -- e.g. statistical dialog act prediction -- require a repair
processing step to resolve contradictions like the one in fig. \ref{segmentation}.

\begin{figure}[htb]
\begin{center}
\mbox{\epsfxsize=8cm
        \epsffile{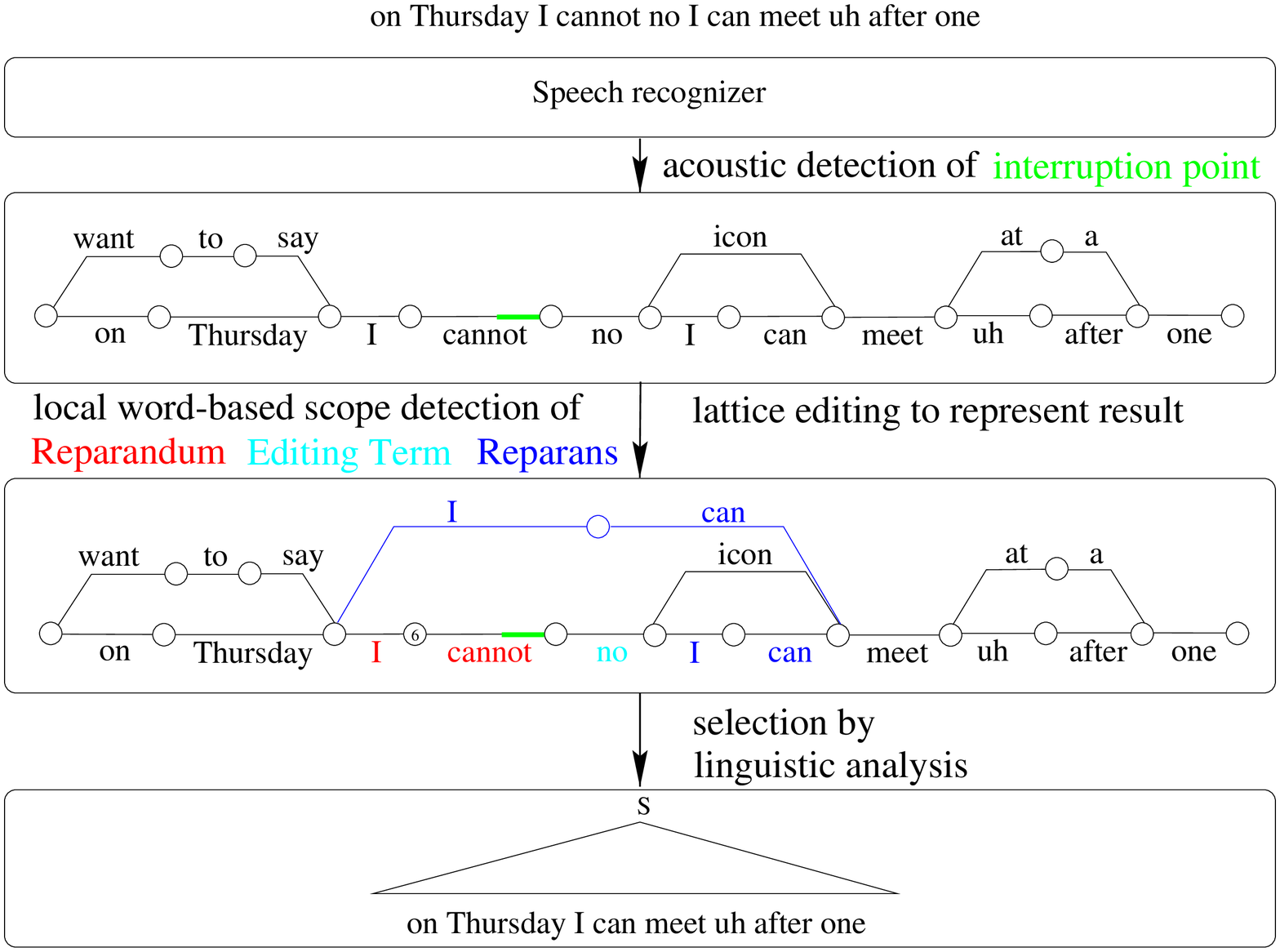}} 
\end{center}
\vspace*{-.65cm}
\caption{\label{architecture}An architecture for repair processing}
\end{figure}

We propose an algorithm for word lattices that divides repair
detection and correction in three steps (cf. fig. \ref{architecture})
First, a trigger indicates potential IPs. Second, a stochastic
model tries to find an appropriate repair for each IP
by guessing the most probable segmentation. To accomplish this, repair processing
is seen as a statistical machine translation problem where the
reparandum is a translation of the reparans. For every repair found, a
 path representing the speaker's intended word sequence is inserted
 into the lattice. In the last step, a lattice parser selects the best path.  \vspace*{-.1cm}
\vspace*{-.2cm}

\section{Repair Triggers}

\label{acoustics}

Because it is impossible for a real time speech system to check for every
word whether it can be part of a repair, we use triggers which
indicate the potential existence of a repair. These triggers must be
immediately detectable for every word in the lattice. Currently we are using two different kind of triggers{\footnote{Other triggers can be added as well. \cite{stolcke:1999} for
example  integrate prosodic cues and an extended language model in a
speech recognizer to detect IPs.}:
\begin{enumerate}
\vspace*{\itspace}
\item Acoustic/prosodic cues:  Speakers mark the IP
  in many cases by prosodic signals like pauses, hesitations, etc.  A
  prosodic classifier
\footnote{The classifier is developed by the
    speech group of  the IMMD 5. Special thanks to Anton Batliner,
    Richard Huber and Volker Warnke.} 
determines for every word the
  probability of an IP following. If it is above a certain threshold, the trigger becomes
  active. For  a detailed description of the acoustic aspects see
  \cite{batliner:1998}.
  \vspace*{\itemspace}
\item Word fragments are a very strong repair indicator. Unfortunately,
  no speech recognizer is able to detect word
  fragments to date. But there are some interesting approaches to
  detect words which are not in the recognizers vocabulary
  \cite{klakow:1999}. A word fragment  is normally an unknown word and
  we hope that it can be distinguished from unfragmented unknown words by
  the prosodic classifier. So, currently this is a
  hypothetical trigger. We will elaborate on it in the evaluation section (cf.
  sect. \ref{results}) to show the impact of this trigger.
\end{enumerate}
  If a trigger is active, a search for an acceptable segmentation into reparandum, editing term and reparans is initiated.

\section{Scope Detection}

As mentioned in the introduction  repair segmentation is based mainly
on a stochastic translation model. Before we explain it in detail we
give a short introduction to statistical machine
translation\footnote{A more detailed introduction is given by
  \cite{brown:1990}}. The fundamental idea is
the assumption that a given sentence $S$ in a source language
(e.g. English) can be translated in any sentence $\hat{T}$ in a target
language (e.g. German). To every pair $(S,\hat{T})$ a probability is
assigned which reflects the likelihood that a translator who sees $S$
will produce $\hat{T}$ as the translation. The statistical machine translation problem is formulated as:
\vspace*{\eqspace}
\begin{equation} \hat{T}= argmax_{T} P(T|S)\vspace*{\eqendspace} \end{equation}
This is reformulated by Bayes' law for a better search space reduction,
but we are only interested in the conditional
probability $P(T|S)$. For further processing steps we have to introduce the
concept of alignment \cite{brown:1990}. Let $S$ be the word sequence
$S_1,S_2 \dots S_l \equiv S_1^l$ and $T= T_1,T_2 \dots T_m \equiv
T_1^m$. We can link a word in $T$ to a word in $S$. This reflects the assumption that the word in $T$ is translated from the word
in $S$. For example, if $S$ is ``On Thursday'' and $T$ is ``Am
Donnerstag'' ``Am'' can be linked to ``On'' but also to
``Thursday''. If  each word in $T$ is linked to exactly one word in $S$
these links can be described by a vector $a_1^m=a_1 \dots a_m$ with $a_i \in {0
\dots l}$. If the word $T_j$ is linked to $S_i$ then $a_j=i$. If it is not connected to any word
in $S$ then $a_j=0$. Such a vector is called an alignment
$a$. $P(T|S)$ can now be expressed by 
\vspace*{\eqspace}
\begin{equation} P(T|S) = \sum_{\mbox{a is alignment}} P(T,a|S)
  \label{sum} \vspace*{\eqendspace} \end{equation}
Without any further assumptions we can infer the following:
\vspace*{\eqspace}
\begin{eqnarray} P(T,a|S) & = & P(m|S) * \nonumber \\
  && \prod_{j=1}^{m} P(a_j|a_1^{j-1},{T}_1^{j-1},m,S)*\nonumber \\[-0.25cm]
& & \hspace*{+0.6cm} P(T_j|a_1^j,{T}_1^{j-1},m,S)
\label{fundamental_eq}
\vspace*{\eqendspace}
\end{eqnarray}
Now we return to self corrections. How can this framework help to
detect the segments of a repair? Assume we have a lattice path where
the reparandum ($RD$) and the reparans($RS$) are given, then $(RS,RD)$ can
be seen as a translation pair and $P(RD|RS)$ can be expressed exactly the
same way as in equation (\ref{sum}). Hence we have a method to
score $(RS,RD)$ pairs. But the triggers only indicate the interruption
point, not the complete segmentation. Let us first look at editing
terms. We assume  them to be a closed list of short phrases. Thus if  an entry
of the editing term list is found after an
IP, the corresponding words are skipped. Any subsequence of
words before/after the IP could be the reparandum/reparans. Because turns can
have an arbitrary length it is impossible to compute $P(RD|RS)$ for every
$(RS,RD)$ pair. But this is not necessary at all, if repairs are considered as local phenomena. We restrict our search to a window
of four words before and after the IP. A corpus
analysis showed that 98\% of all repairs are within this window. Now we only
have to compute  probabilities for $4^2$ different pairs.  If the probability of a
$(RS,RD)$ pair is above a certain threshold, the segmentation is accepted as a repair. 

\subsection{Parameter Estimation}
The conditional probabilities in equation (\ref{fundamental_eq}) cannot 
be estimated reliably from any corpus of realistic size, because there
are too many parameters. For example both $P$ in the product depend on the complete
reparans $RS$. Therefore we simplify the probabilities by assuming
that $m$ depends only on $l$, $a_j$ only on $j$,$m$
and $l$ and finally $RD_j$ on $RS_{a_j}$. So equation
(\ref{fundamental_eq}) becomes
\vspace*{\eqspace}
\begin{eqnarray}
\lefteqn{P(RD,a|RS)= P(m|l)*} \nonumber \\
&& \prod_{j=1}^{m} P(a_j|j,m,l)*P(RD_j|RS_{a_j})
\label{fundamental_rep2}
\vspace*{\eqendspace}
\end{eqnarray}
These probabilities can  be directly trained from a manually annotated
corpus, where all repairs are labeled with begin, end, IP and editing term and for each reparandum the words are linked to
the corresponding words in the respective reparans. All distributions are smoothed by a simple
back-off method \cite{katz:1987} to avoid zero probabilities with the
exception that the word replacement probability $P(RD_j|RS_{a_j})$ 
is smoothed in a more sophisticated way.

\subsection{Smoothing}
Even if we reduce the number of parameters for the word replacement probability by
the simplifications mentioned above there are a lot of parameters
left. With a vocabulary size of 2500 words, $2500^2$ parameters have
to be estimated for $P(RD_j|RS_{a_j})$. The corpus\footnote{$\sim$11000turns with $\sim$240000 words} contains  3200 repairs from which we extract about 5000 word links. So
most of the possible word links never occur in the corpus. Some of them are  more likely to occur in a repair than others. For example, the
replacement of ``Thursday'' by ``Friday'' is supposed to be more likely than by
``eating'', even if both replacements are not in the training
corpus. Of course, this is related to the fact that a repair is a
syntactic and/or semantic anomaly. We make use of it by adding
two additional knowledge sources to our model. Minimal syntactic
information is given by part-of-speech (POS) tags and POS sequences,
semantic information is given by semantic word classes. Hence the input is
 not merely a sequence of words but a sequence of triples. Each triple
 has three slots (word, POS tag, semantic class). In the next section we will describe how we obtain these two information pieces for every word in the
 lattice. With this additional information, $P(RD_j|RS_{a_j})$
 probability could be smoothed by linear interpolation of word, POS and
 semantic class replacement probabilities. 
\vspace*{\eqspace}
\begin{eqnarray}
\lefteqn{P(RD_j|RS_{a_j}) = } \nonumber \\
&& \alpha*P(Word(RD_j)|Word(RS_{a_j}))\nonumber \\
&&{}+\beta*P(SemClass(RD_j)|SemClass(RS_{a_j}))\nonumber \\
&&{}+\gamma*P(POS(RD_j)|POS(RS_{a_j}))
\end{eqnarray}
 with $ \alpha+\beta+\gamma=1 $.

$Word(RD_j)$ is the notation for the selector of the word slot of the
triple at position $j$.

\section{Integration with Lattice Processing}
We can now detect and correct a repair, given a sentence annotated with
POS tags and semantic classes. But how can we construct such a sequence
from a word lattice?
Integrating the model in a lattice algorithm requires three steps:
\begin{itemize}
\vspace*{\itspace}
\item mapping the word lattice to a tag lattice \vspace*{\itemspace}
\item triggering IPs and extracting the possible reparandum/reparans pairs \vspace*{\itemspace}
\item introducing new paths to represent the plausible reparans
\end{itemize}
\vspace*{\itendspace}
The tag lattice construction is adapted from \cite{samuelsson:1997}. For
every word edge and every denoted POS tag a corresponding tag 
edge is created and the resulting probability is determined. If a tag edge already
exists, the probabilities of both edges are merged.  The original
words are stored together with their unique semantic class in a
associated list.
Paths through the tag graph are scored by
a POS-trigram. If a trigger is active, all paths through the word before
the IP need to be tested whether an acceptable repair
segmentation exists. 
Since the scope model takes at most four words for reparandum and
reparans in account it is sufficient to expand only partial
paths. Each of these partial paths is then processed by the scope
model. To reduce the search space, paths with a low score can be pruned.

Repair processing is integrated into the Verbmobil system as a filter
process between speech recognition and syntactic analysis. This
enforces a repair representation that can be integrated into a
lattice. It is not possible to mark only the words with some
additional information, because a repair is a phenomenon that depends
on a path. Imagine that the system has detected a repair on a certain
path in the lattice and marked all words by their repair function. Then
a search process (e.g. the parser) selects a different 
path which shares only the words of the reparandum. But these words
are no reparandum for this path. A solution is to introduce a new path
in the lattice where reparandum and editing terms are deleted. As we
said before, we do not want to delete these segments, so they are stored
in a special slot of the first word of the reparans. The original
path can now be reconstruct if necessary.

To ensure that these new paths are comparable to other paths we
score the reparandum the same way the parser does, and add the resulting value
to the first word of the reparans. As a result, both the original path and
the one with the repair get the same score except one word transition. The 
(probably bad) transition in the original path from the last word of
the reparandum to the first word of the reparans is replaced by a
(probably good) transition from the reparandum's onset to the
reparans. We take the lattice in fig. \ref{architecture} to give an
example. The scope model has marked ``I cannot'' as the reparandum,
``no'' as an editing term, and ``I can'' as the reparans. We sum up the
acoustic scores of ``I'', ''can'' and ``no''. Then we add the maximum
language model scores for the transition to ``I'', to ``can'' given ``I'',  and to
``no'' given ``I'' and ``can''. This score is added as an offset to
the acoustic score of the second ``I''.

\section{Results and Further Work}
\label{results}
Due to the different trigger situations  we performed two tests: One
 where we use only acoustic triggers and another where the existence of a perfect word fragment detector is assumed.  The input were unsegmented transliterated utterance to
exclude influences a word recognizer. We restrict the processing time
on a SUN/ULTRA 300MHZ to 10 seconds. The parser was simulated by a
word trigram. Training and testing were done on two separated parts of
the German part of the Verbmobil corpus (12558 turns training / 1737 turns test). 
\begin{table}[htb]
\centering
\small
\begin{tabular}{|r|r|r|r|r|}\hline
 & \multicolumn{2}{|c|}{Detection}
& \multicolumn{2}{|c|}{Correct scope} \\ \hline
& \small Recall & \small Precision & \small Recall & \small Precision \\ \hline
Test 1 & 49\% &  70\% & 47 \% & 70\% \\ \hline
Test 2 & 71\% & 85\% & 62\% & 83\% \\ \hline
\end{tabular}
\end{table}

A direct comparison to other groups  
is rather difficult due to very different corpora, evaluation
conditions and goals. \cite{nakatani:1993}
suggest a acoustic/prosodic detector to identify IPs
but don't discuss the problem of finding the correct
segmentation in depth. Also their results are obtained on a corpus
where every utterance contains at least one repair. \cite{shriberg:1994} also
addresses the acoustic aspects of repairs. Parsing approaches
like in \cite{bear:1992,hindle:1983,core:1999} must be proved to work
with lattices rather than  transliterated text. An algorithm which is
inherently capable of lattice processing is proposed by Heeman \cite{heeman:1997b}. He redefines the word
recognition problem  to identify the best sequence of words,
corresponding POS tags and special repair tags. He reports a recall rate of 81\% and a precision of 83\% for
detection and 78\%/80\% for correction. The test settings are nearly
the same as test 2. Unfortunately, nothing is said about the processing time of his
module. 

We have presented an approach to score
potential reparandum/reparans pairs with a relative simple scope
model. Our results show that repair processing with statistical
methods and without deep syntactic knowledge is a promising approach
at least for modification repairs. Within this framework  more sophisticated scope
models can be evaluated. A system integration as a
filter process is described. Mapping the word lattice to a POS tag
lattice is not optimal, because word information is lost in the
search for partial paths. We plan to implement a combined combined POS/word tagger. 
\bibliography{diss_eng}

